\documentclass[sigconf]{acmart}
\usepackage{array}
\usepackage{graphicx}
\usepackage{booktabs,tabularx,ragged2e}
\newcolumntype{L}[1]{>{\RaggedRight\arraybackslash}p{#1}}

\AtBeginDocument{%
}

\copyrightyear{2025}
\acmYear{2025}
\setcopyright{acmlicensed}\acmConference[SA Art Papers '25]{SIGGRAPH Asia 2025 Art Papers}{December 15--18, 2025}{Hong Kong, Hong Kong}
\acmBooktitle{SIGGRAPH Asia 2025 Art Papers (SA Art Papers '25), December 15--18, 2025, Hong Kong, Hong Kong}
\acmDOI{10.1145/3757369.3767602}
\acmISBN{979-8-4007-2129-8/2025/12}

\begin{document}
\citestyle{acmauthoryear}

\title{Airy: Reading Robot Intent through Height and Sky}

\author{Baoyang Chen}
\orcid{0009-0009-4882-6788}
\email{baoyang.chen@gmail.com}
\affiliation{%
  \institution{Hong Kong University of Science and Technology}
  \city{Hong Kong SAR}
  \country{China}
}
\affiliation{%
  \institution{Central Academy of Fine Arts}
  \city{Beijing}
  \country{China}
}

\author{Xian Xu}
\orcid{0000-0002-2636-7498}
\email{xianxu@ln.edu.hk}
\affiliation{%
  \institution{Lingnan University}
  \city{Hong Kong SAR}
  \country{China}
}

\author{Huamin Qu}
\orcid{0000-0002-3344-9694}
\authornote{Corresponding author}
\email{huamin@ust.hk}
\affiliation{%
  \institution{Hong Kong University of Science and Technology}
  \city{Hong Kong SAR}
  \country{China}
}

\begin{abstract}
As industrial robots move into shared human spaces, their opaque decision making threatens safety, trust, and public oversight. This artwork, \textit{Airy}, asks whether complex multi agent AI can become intuitively understandable by staging a competition between two reinforcement trained robot arms that snap a bedsheet skyward. Building on three design principles-competition as a clear metric (“who lifts higher?”), embodied familiarity (audiences recognize fabric snapping), and sensor to sense mapping (robot cooperation or rivalry shown through forest and weather projections)-the installation gives viewers a visceral way to read machine intent. Observations from five international exhibitions indicate that audiences consistently read the robots’ strategies, conflict, and cooperation in real time, with emotional reactions that mirror the system’s internal state. The project shows how sensory metaphors can turn a black box into a public interface.
\end{abstract}



\begin{teaserfigure}
  \centering
  \includegraphics[width=\textwidth]{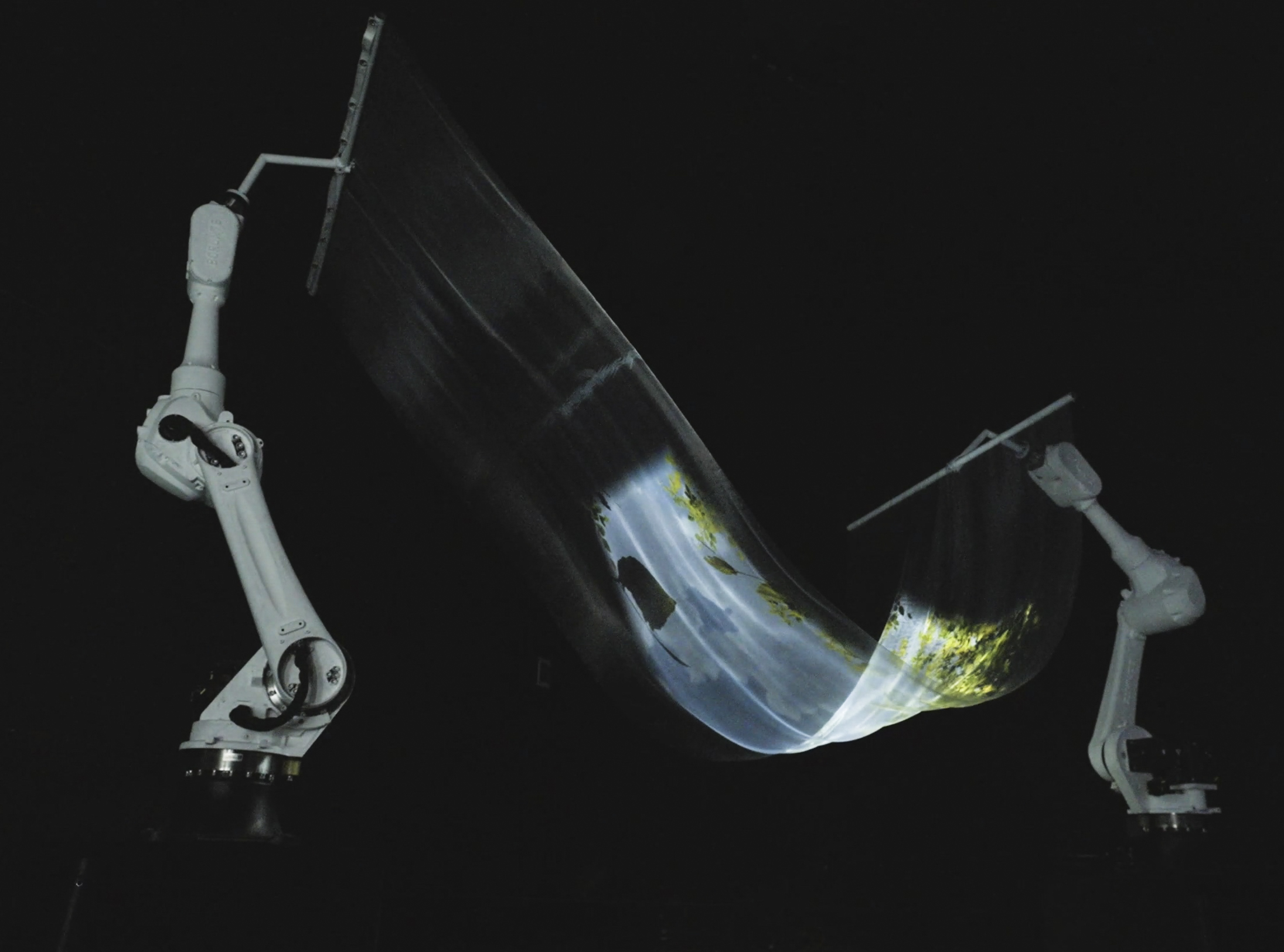}
  \Description{Two white industrial arms face each other holding a lightweight silk veil. A projected forest scene is visible on the veil; an audience stands at a safe distance.}
  \caption{Installation Overview of the Artwork}
  \label{fig:teaser}
\end{teaserfigure}

\maketitle

\section{Introduction}

Industrial manipulators assist surgeons, greet museum visitors, and roam warehouses, yet observers still struggle to \emph{read} what these embodied algorithms will do next \citep{Breazeal2003,DraganLeeSrinivasa2013}. When legibility fails, even advanced control becomes an opaque, and potentially unsafe, black box.

Media artists confronted this opacity long before domestic service robots. Early cybernetic sculptures, Schöffer’s \emph{CYSP~1} and Ihnatowicz’s \emph{Senster} \citep{Schoffer1956,Ihnatowicz1970}, made feedback a public spectacle. Networked works such as Goldberg and Santarromana’s \emph{Telegarden} \citep{GoldbergSantarromanaBekey1995} and Stelarc’s body extensions \citep{Stelarc1981} redistributed agency among artist, machine, and audience \citep{Kac2001,Penny2013}. Recent work highlights improvisation and provocation as strategies in artist–robot collaborations \citep{BenfordGarrettSchneiders2024}. Surveys show that contemporary robotic art extends this path across drawing, theatre, music, and dance \citep{Jeon2017}.

A decisive pivot treated \emph{emergent behavior} as material. Work on emergent and self‑organizing systems shows that simple sensorimotor couplings can produce rich, legible structure \citep{Braitenberg1984,DeLanda2011}. Installations that foreground process and conflict—\emph{Sisyphus} \citep{Chan2024}, or self‑assembly, like Dean, D’Andrea, and Donovan’s \emph{Robotic~Chair} \citep{DeanAndreaDonovan2006}, make internal machine constraints publicly legible.

Competition alone does not ensure comprehension. Studies in kinesthetic empathy suggest that audiences can experience movement through embodied resonance \citep{Ooms2023}. Art and science collaborations in robotic drawing document how embodiment can make process visible \citep{TressetLeymarie2013}. Many installations stabilize interpretation by translating core robot signals into ambient environmental metaphors \citep{ChadalavadaAndreassonKrugLilienthal2015}, but such mappings are trusted only when instrumentation is transparent. Cybernetic precursors insisted on visible feedback paths \citep{Pask1968}; projects in robotic theatre and HRI examine audience response and creative agency \citep{PetrovicKicinbaciPetricKovacic2019,SandovalSosaCappuccioBednarz2022}.

Across these strands, scholars describe an emerging literacy: the capacity of non experts to read intention, error, and adaptation in real time. Historic precedents, algorithmic contests, kinesthetic resonance, narrative visualization, and instrumented fairness define the terrain shared by media art and HRI. Our research questions are: \textbf{Can a single competitive metric (height), paired with a familiar domestic gesture (sheet snapping) and a truthful sensor to sense mapping (weather/forest), render multi agent control intent legible to lay audiences in real time?}

\section{Motivation and Concept}
\label{sec:motivation}

Algorithms already move markets, filter headlines, and match our dates; the next frontier is \emph{embodied} AI sharing our streets, homes, and hospitals. \textbf{How can we interpret their embodied behaviors, and how can an artist harness machine emergence as material?} Answering requires both legibility and creative control, so we adopt a triadic lens that distributes authorship among \emph{artist}, \emph{machine}, and \emph{audience}, each providing distinct insight. To balance these agencies, spectators need rapid, intuitive \emph{emergence literacy}: the ability to read intention, error, and adaptation in real time.

We use competition as a universal decoder: as in sport, a single metric, ``who gets higher?'', turns invisible optimization into a visible race. Framing the installation as a height tug between two robotic arms lets viewers read strategy much like a high jump final, rising arcs, near misses, and the crowd’s roar or groan.

For embodied familiarity, we elevate a mundane chore, snapping a bedsheet, that already lives in muscle memory. The familiar dip and whip that billows fabric gives the body an internal physics model that predicts what should happen next. Giving the same task to two robot arms turns spectators into empathetic commentators who can \emph{feel}, before they see, whether an upswing will succeed.

\begin{figure*}[t]
  \centering
  \includegraphics[width=\linewidth]{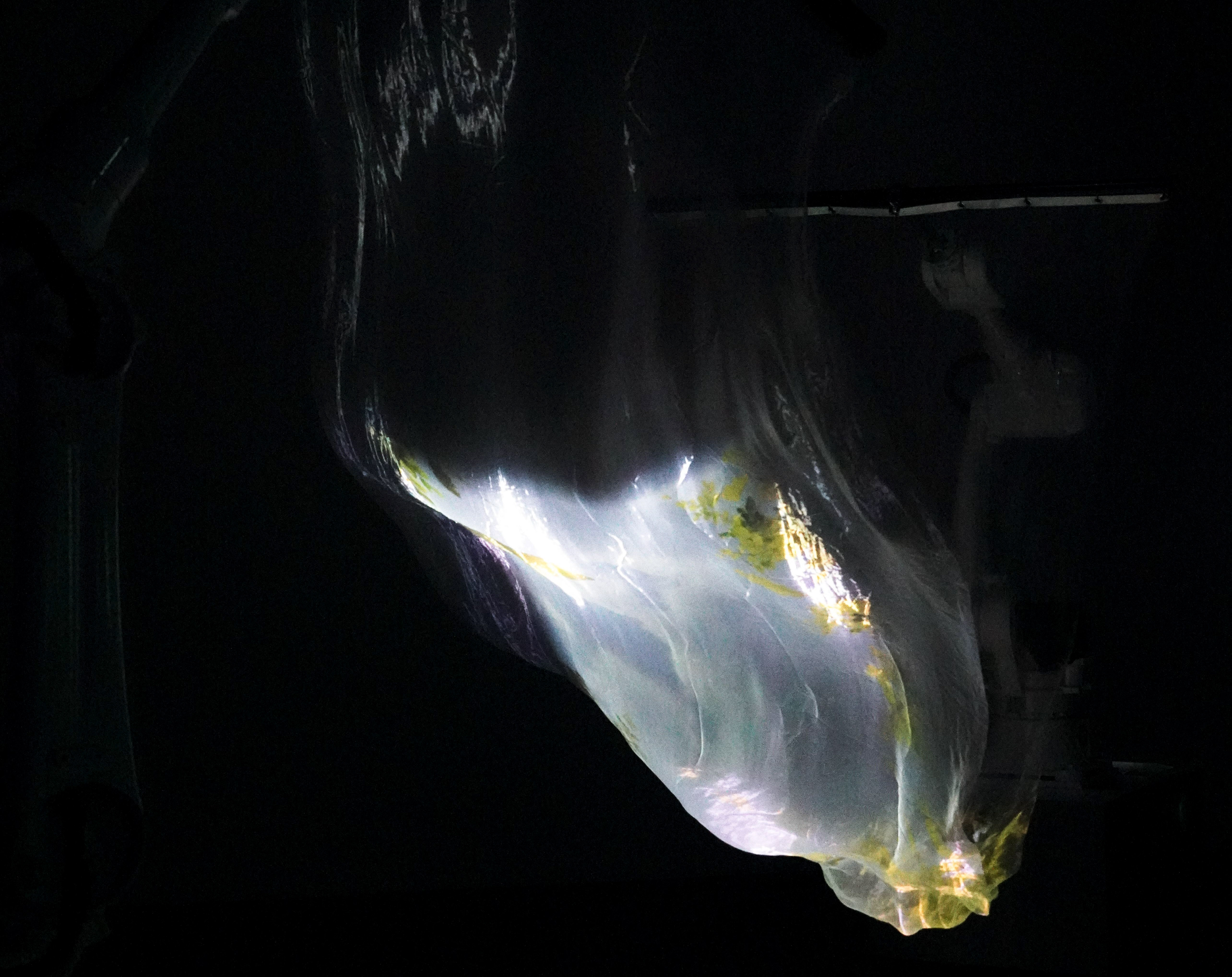}
  \caption{A forest slice projected onto the lightweight silk.}
  \Description{A forest slice projected onto the lightweight silk floating in the air.}
  \label{fig:instl1}
\end{figure*}

We add a visual altimeter. Instead of numeric readouts, the silk itself carries a forest panorama, ferns, trunks, canopy, then sky, so each centimeter of height \emph{unlocks} a rarer layer and a clearer view, making ``higher'' instantly legible as ``better.''

Because people read blue skies as harmony and storms as trouble, we implement an atmospheric visualizer. Each arm’s controller continually weighs cooperation versus competition; tinting the projected sky exposes that negotiation: tranquil blue when both work together, turbulence and lightning when one seizes the lead. Weather becomes a live scoreboard.

True competition needs a fair judge. The installation measures both \emph{absolute} veil height and each arm’s \emph{relative} energy input. Without objective signals, neither robots nor spectators can close the perception action loop. Section~\ref{sec:system} details the sensing hardware and data flow; here we focus on \emph{why} clarity matters.

The work is driven by three interlocking forms of agency, distinct, mutually dependent, and each supplying information the others lack:
\begin{description}
  \item[Artist: The World Builder.]
  Sets constraints that make the contest legible: selects ultralight silk that amplifies every nudge; couples altitude to a scrolling forest panorama and timing error to dynamic weather so raw numbers become sights and sounds; and tunes the reinforcement signal, credit for new height, cost for wasted motion, so risk, cooperation, and rivalry stay in balance.
  \item[Robots: The Strategic Actors.]
  Within those rules, each arm runs a learned policy, sharing only instantaneous joint angles (never future plans). Every lift is both move and bluff. By adjusting phase and force, they sometimes ally for altitude and sometimes defect for a lead, turning visible choreography into the trace of an invisible negotiation.
  \item[Spectators: The Sense Makers.]
  Viewers close the loop by turning motion into meaning: because sheet snapping sits in muscle memory, they sense success or failure a split second early, and read weather shifts, clear versus storm, as cues of harmony or conflict.
\end{description}

Rather than a polished AI output, we stage a public laboratory where algorithmic motives are legible through bodies, landscape, and weather. If two machines can declare intent on a sheet of silk, the same principle can scale to larger algorithmic arenas that remain black boxes.

\section{Artwork Appearance}
\label{sec:appearance}

\begin{figure*}[t]
  \centering
  \includegraphics[width=\linewidth]{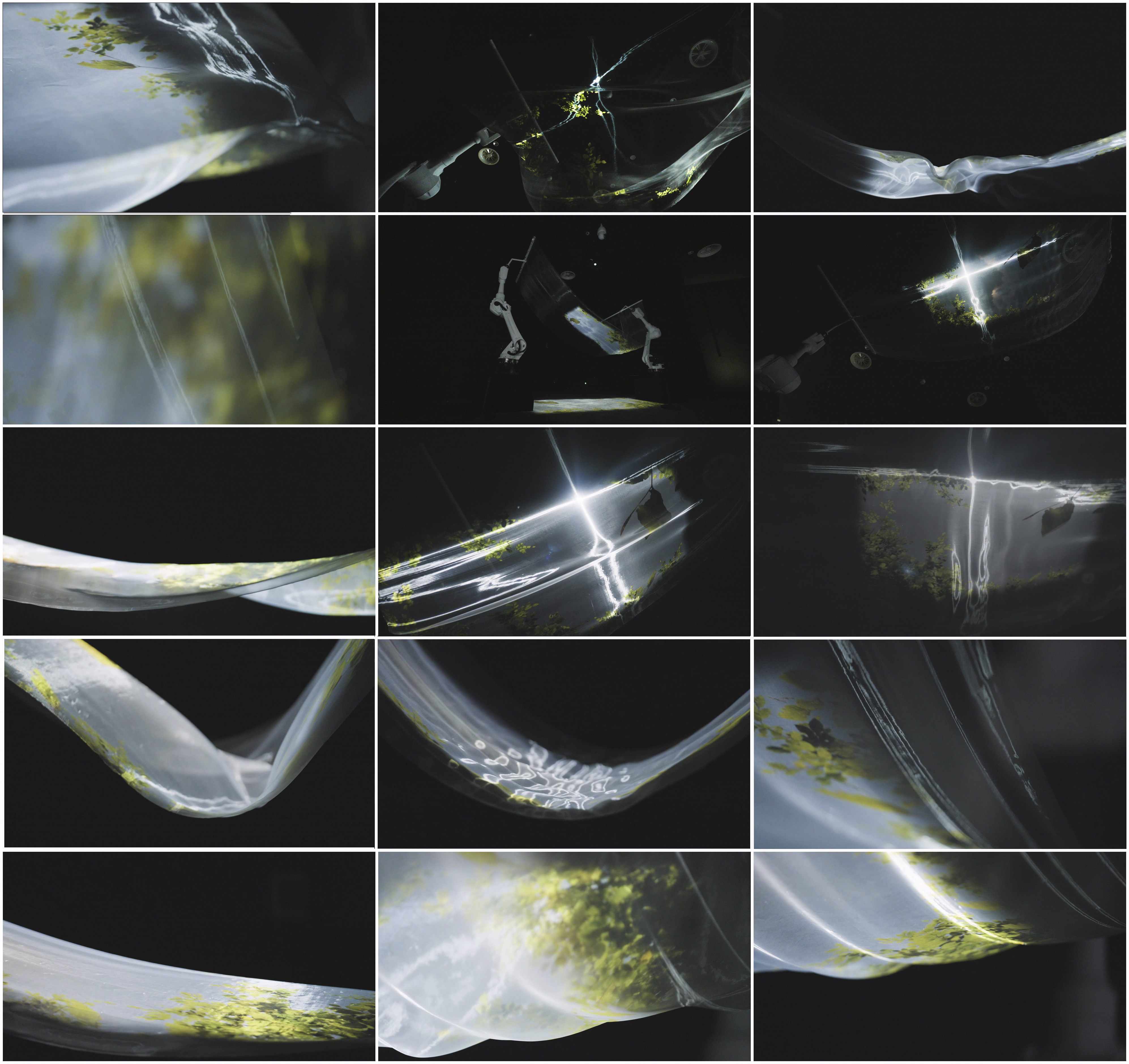}
  \caption{Stills from the documentation video highlighting motion.}
  \Description{Sixteen-panel contact sheet. Across close-ups and two wide shots, a translucent silk veil spans between two white industrial arms in a dark gallery. Frames alternate between (i) micro views of the veil’s catenary curve, ripples, and edge hem, (ii) mid-shots where projected forest imagery—moss, trunks, canopy, patches of blue sky—scrolls along the fabric, and (iii) moments of atmospheric presets tied to timing: soft overcast, mist, and a bright lightning strike rendered as a sharp vertical flare. Motion blur traces snap lifts and phase shifts; the veil moves from slack troughs to high crests. Arm end-effectors, ceiling fixtures, and projector highlights are visible; no spectators appear. The sequence evidences the mapping height to forest altitude and timing relation to weather state.}
  \label{fig:instl2}
\end{figure*}

\textit{Airy} inhabits a darkened room where two white industrial robot arms face each other across an ultralight silk sheet. The fabric is an interface, not a backdrop: small arm motions inject energy into the veil, raising continuous micro ripples that make every competitive adjustment visible.

A high resolution forest cross section is projected onto the translucent silk. As the fabric rises or falls, the image scrolls in sync, as if the silk were a living window. Near the floor it settles into a mossy understory with grass and scattered pinecones; mid height brings sun dappled trunks and swaying ferns; higher still, dense foliage; at the peak, the silk turns nearly weightless and translucent, opening a patch of sky framed by treetops.

As detailed in the atmospheric visualizer rationale (\S\ref{sec:motivation}), a generative weather layer mirrors moment to moment rapport between the arms. Collaboration bathes the forest in tranquil light; a delay or surge that signals rivalry flips the system into storm logic.

\section{System Design}
\label{sec:system}
\begin{figure}[t]
  \centering
  \includegraphics[width=0.5\textwidth]{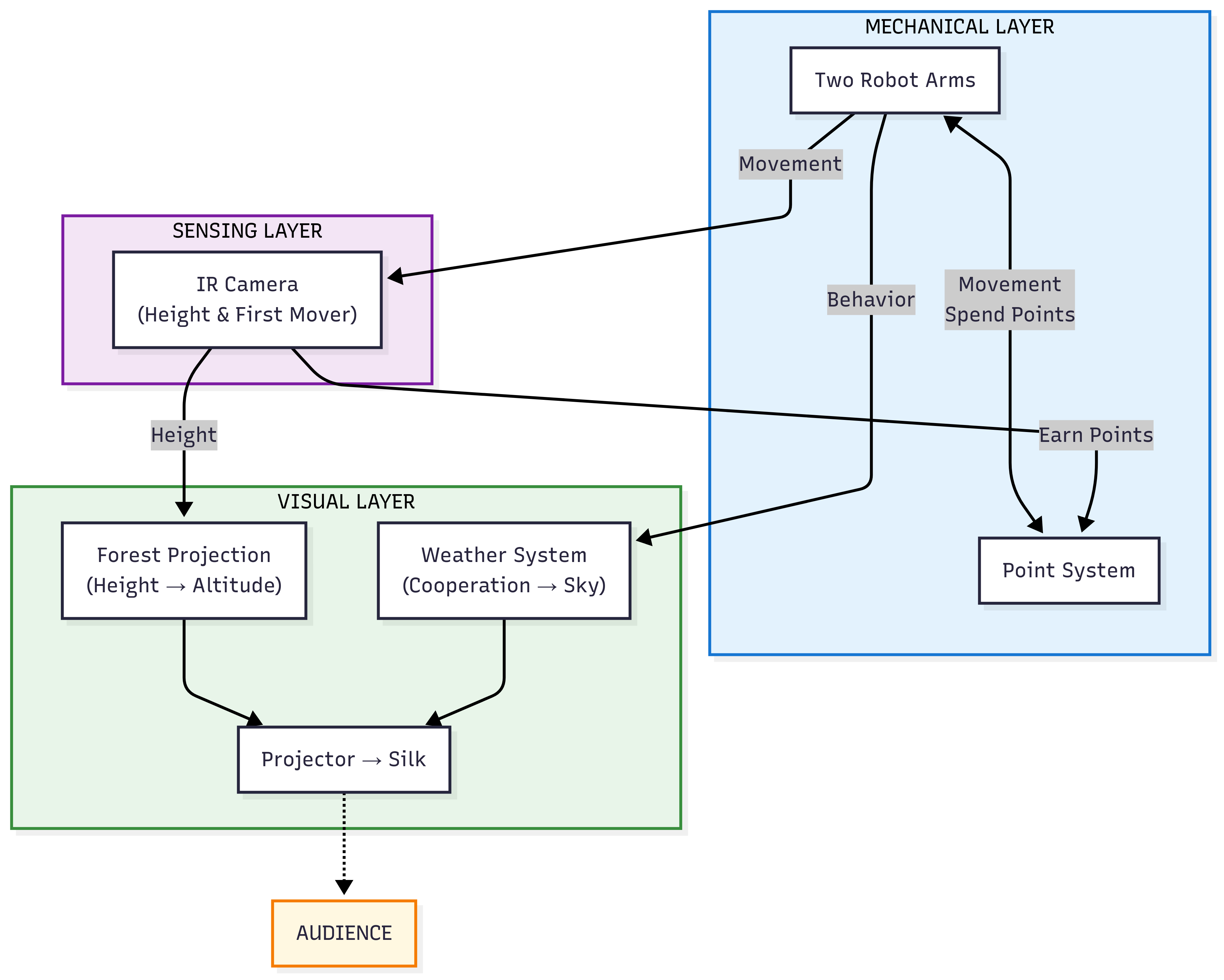}
  \caption{System layout overview.}
  \label{fig:layout}
\end{figure}

The work is a layered conversation: every voice (robot, silk, projection, weather, sound, audience) is both speaker and listener. Figure~\ref{fig:layout} sketches the information flow; the layers below outline each role concisely. In overview, (i) When the arms move in step, the silk rises cleanly and the system selects the clear sky preset. (ii) When their timing splits, ascent stalls or dips and the system selects a storm preset. (iii) When safety triggers, motion softens immediately and the scene falls to a blue hush.

\subsection*{Mechanical Layer: The Duelists}
Two six axis industrial arms grip the short edges of an ultralight silk sheet. Each runs its own policy with a single objective: lift as high as possible without tearing. Scoring is simple:
\begin{itemize}
  \item \textbf{Earn credit} only by setting a new height record.
  \item \textbf{Spend credit} on every motion, large or small.
\end{itemize}
Because fidgeting drains the score, the best strategy is to wait, then launch decisive moves.

\subsection*{Sensing Layer: The Referee}
The sensing layer supplies three objective signals—peak height, first mover, and timing relation—that drive the control loop and public cues. Peak height is computed by an overhead infrared camera that estimates the silk’s center height; beating the rolling 180\,s record triggers a score. The same camera identifies the \emph{first mover}, tagging which arm initiated a record-breaking lift (credit is shared if starts are near-simultaneous). A lightweight estimator derives the \emph{timing relation} from joint encoders by comparing the arms’ edge velocities and classifying them as \emph{in-step}, \emph{small lag}, \emph{growing lag}, or \emph{split}; this cue informs both the control policy and the weather logic.

\subsection*{Visual Layer: Projection}
This layer translates motion into an intuitive landscape narrative. A high lumen projector maps a seamless forest panorama, fern floor, trunk zone, canopy, open sky, onto the silk. A virtual camera tied to the silk’s center scrolls upward with each lift, revealing higher ecological layers as if climbing. A compact ruleset turns the \emph{timing relation} into sky changes. When both arms move in‑step, the cooperation cue is high; solo surges lower it. That single cue selects one of four weather presets (hysteresis prevents flicker):

\begin{table}[h!]
\centering
\caption{Weather presets are driven by timing cues.}
\begin{tabular}{@{}lll@{}}
\toprule
\textbf{Arm behavior} & \textbf{Visual weather} & \textbf{Audience feel} \\
\midrule
Cooperative / steady & Clear sun, drifting pollen & Easy harmony \\
Small phase lag & Light overcast, soft wind & Quiet tension \\
Growing lag or jitter & Mist, distant thunder & Rising strain \\
Sudden split & Lightning flash, hard rain & Open conflict \\
\bottomrule
\end{tabular}

\label{tab:weather}
\end{table}

\begin{table*}[ht!]
\centering
\caption{Signals, sources, and how they inform control and public cues.}
\label{tab:signal_path}
\footnotesize
\begin{tabularx}{\linewidth}{@{}L{2.8cm} L{2.6cm} L{3.2cm} L{2.8cm} L{1.6cm}@{}}
\toprule
\textbf{Signal} & \textbf{Source} & \textbf{Controller} & \textbf{Visuals} & \textbf{Safety} \\
\midrule
Silk peak height & Overhead camera & Yes (lift \& scoring) & Forest altimeter & Indirect \\
Timing relation (in-step / drift / split) & Encoder-based estimator & Yes (phase decisions) & Weather preset & Indirect \\
Safety flag (tension or torque spike) & Joint sensors & Yes (soften / pause) & Blue hush & Direct trigger \\
First mover & Camera timing check & Yes (credit logic) & No & None \\
\bottomrule
\end{tabularx}
\end{table*}

\subsection*{Decision Layer: How the Robots ``Decide''}
Each arm chooses \emph{human‑nameable} motion ingredients, \emph{how large to lift, when to time the snap, when to dwell}, which the native controller renders as smooth, jerk‑limited wrist and shoulder trajectories. The result is musical phrasing rather than jitter. Energy leaves the grippers as a traveling wave in the silk; when waves meet in time, the sheet rises in one clean breath. Cooperation emerges through timing: in‑step motion is amplified; misaligned surges prompt a yield or a counter. This is where rivalry appears, and spectators begin to narrate motive.

\subsection*{Algorithmic System: Learning \& Control}
Each arm executes a frozen RL policy trained in a cloth simulator with domain randomization (mass, damping, friction). The policy observes only present-tense signals needed for legible action: the silk’s current height and trend (camera), the arm’s own joint state and a simple tension proxy, and the partner’s \emph{current} pose (never its future plan). Rather than output torques, the policy selects a compact motion primitive lift amplitude, snap phase (relative to the last detected wave crest), and dwell, which the native stack converts to jerk-limited joint trajectories under impedance control with strict torque/velocity limits. Training rewards mirror the scoring (credit for new peaks, cost for motion, strong penalties near safety limits). At runtime, policies are fixed; a watchdog triggers the blue-hush failsafe on tension/torque spikes; and the arms exchange only their \emph{current} poses at a modest rate, so cooperation or rivalry emerges from timing rather than negotiation.

\subsection*{One Second On Stage}
Each second the loop closes: the camera updates height; the timing estimator updates phase; each arm feels its joints and the cloth’s give; the decision layer either breathes with the fabric or waits; motors nudge or hold; the projection lifts the forest window and either clears the sky or teases in cloud. Logs capture these micro‑choices so we can align what the audience felt with what the robots did.


\begin{table*}[ht!]
\centering
\caption{Hypothesis mapped to internal signals, public cues, and episodes.}
\label{tab:hypothesis_alignment}
\begin{tabular}{@{}llll@{}}
\toprule
\textbf{Hypothesis part} & \textbf{Internal signals} & \textbf{Public cues} & \textbf{Episode} \\
\midrule
(i) cooperation legible & in-step timing; rising height & smooth ascent; clear sky & Clear Ascent \\
(i) cooperation legible & small timing drift & tremor near canopy; overcast & Suspended Negotiation \\
(ii) conflict legible & timing split; stall or dip & storm preset; turbulence & Competitive Whiplash \\
(i) cooperation legible & timing realignment & steady rise; sky clears & Recovery Spiral \\
(iii) safety legible & safety flag; torque spike & softened motion; blue hush & Safety Eclipse \\
\bottomrule
\end{tabular}
\end{table*}

\subsection*{Safety Reflexes and Honest Visuals}
A quiet referee watches for trouble. If tension or torque spikes, the arms soften instantly and the scene falls to a blue hush safety over victory. The same signals that guide the arms also color the room: real lift raises the forest; in‑step motion keeps light clear; mist and gusts arrive when timing drifts. Because visuals are driven by the robots’ own cues, weather reads as explanation, not decoration.

\section{Results And Discussion}
\label{sec:results}

Debuting at a major international biennial in summer 2024 and later touring leading museums, the installation has logged tens of thousands of interactions. Field observations ground the discussion. Because veil height maps to vertical parallax and timing error to atmospheric entropy, spectators infer intent without code.

\subsection*{Trace Analysis Without Participants}
The system logs height estimates, per arm timing cues, controller states, weather preset changes, and safety flags. We identify five recurring moments already in the piece, \textit{Clear Ascent, Suspended Negotiation, Competitive Whiplash, Recovery Spiral}, and \textit{Safety Eclipse}, directly from these traces (e.g., a storm switch following a surge in timing mismatch) and align them to the projection state. We use these moments to test the hypothesis in plain language.

\subsection*{Thematic Episodes Aligned to the Research Hypothesis}

\paragraph{Clear Ascent.}
When both arms fire their learned snap lift in perfect counter phase, the veil sails upward and the forest opens to cloudless sky. Warm light floods the silk, pollen drifts, and the servo chorus resolves to a single confident note.
\emph{Link to hypothesis:} clean rise and clear sky co-occur with in step motion, as predicted.

\paragraph{Suspended Negotiation.}
At times one wrist probes upward while its partner hovers, testing tension while sharing only present joint states over the limited link. The veil trembles below the canopy, light cools to overcast, and a faint breeze unsettles leaves, the signature of a small phase lag.
\emph{Link to hypothesis:} minor timing drift appears as overcast and micro corrections, signaling emerging disagreement without numbers.

\paragraph{Competitive Whiplash.}
An overshoot meets a forceful counter snap. Roots lurch into view, sky darkens, wind surges, thunder rolls, a vivid index of a large timing split that neither arm can predict, since future commands are never shared.
\emph{Link to hypothesis:} a sudden split stalls ascent and triggers storm, making conflict legible as it happens.

\paragraph{Recovery Spiral.}
After the clash, jittery corrections gradually relock phase. Mist lifts, shafts of light return, and the veil steadies into ascent.
\emph{Link to hypothesis:} as timing realigns, ascent steadies and weather clears, restoring legibility.

\paragraph{Safety Eclipse.}
If torque exceeds the reflex threshold, motion softens and freezes. The silk slumps, projection fades to blue twilight, and a soft rain hiss replaces mechanical song.


\begin{figure}[ht!]
  \centering
  \includegraphics[width=0.5\textwidth]{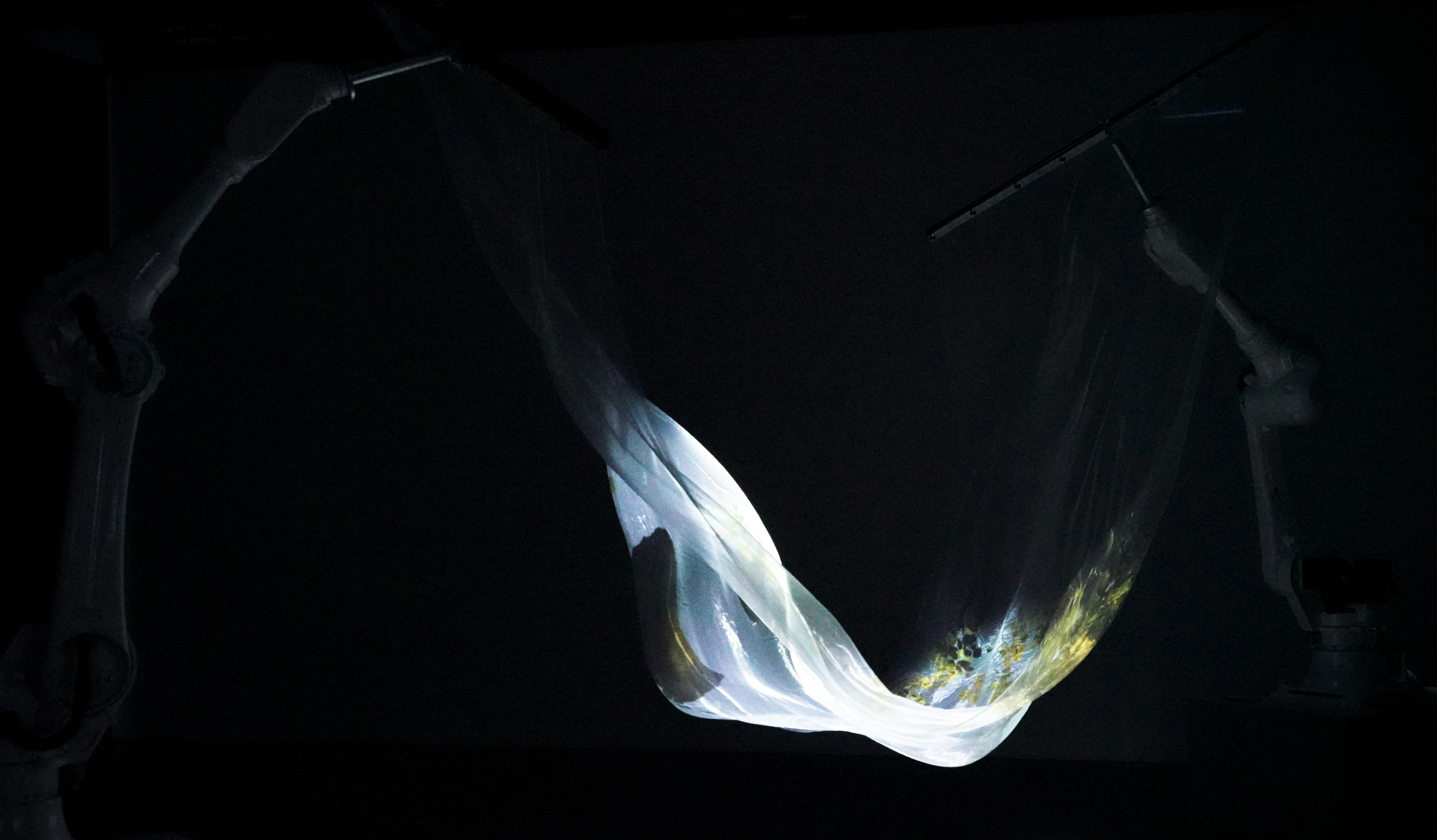}
  \caption{Installation view of the silk floating.}
  \Description{Installation view of two robotic arm swing the silk floating.}
  \label{fig:DSC00847s2}
\end{figure}

Across episodes, the same timing cue that guides control also drives the sky logic. Clear ascent co-occurs with in step timing; storms with timing splits; safety with a visible hush. These co-occurrences match the hypothesis. Our argument relies on internal traces and designed logic rather than audience ratings. A strong counterexample would show sustained clear sky during verified timing splits, or storms during verified harmony. Releasing representative traces enables independent replay to probe such cases.

\section{Conclusion}

This installation tested a simple proposition: if the hidden variables that govern multi agent control are mapped onto sensory channels people already know in their bodies, then algorithmic intent can become publicly legible without spreadsheets or saliency maps. By translating silk height into forest altitude and timing alignment into atmospheric clarity, the work invited visitors to \emph{feel} cooperation and conflict rather than merely read about them. The episode structure and trace alignment provide preliminary support for our guiding claim that a single competitive metric plus embodied familiarity and truthful mapping can make machine strategy legible in real time.

Short video excerpts of the premiere circulated widely online, amassing substantial public attention and commentary across major platforms within six months. The momentum translated into curatorial interest, with multiple international media arts festivals and contemporary art venues requesting touring editions.
Meanwhile, our findings remain provisional, drawn from exhibition contexts. Larger multi site evaluations, counter mapping conditions, and accessibility refinements (including considerations for photosensitive viewers) are planned. Enhanced projection tracking, robustness against mechanical wear, and deeper audience sensing could further stabilize legibility across venues.

\begin{acks}
We thank HKUST VISLAB for funding, guidance, facilities, and feedback. This work was also partially supported by
\grantsponsor{moeprc}{Ministry of Education of the People’s Republic of China}, under grant \grantnum{moeprc}{2024011060546},  and by
\grantsponsor{cafa}{Central Academy of Fine Arts}, under grant \grantnum{cafa}{2025YB004}.

\end{acks}

\end{document}